%% file: emnlp2023.tex
\pdfoutput=1

\documentclass[11pt]{article}

\usepackage[]{EMNLP2023}

\usepackage{times}
\usepackage{latexsym}

\usepackage[T1]{fontenc}

\usepackage[utf8]{inputenc}

\usepackage{microtype}

\usepackage{inconsolata}
\usepackage{graphicx}
\usepackage{booktabs}
\usepackage{tabularx}
\usepackage{multirow}
\usepackage{graphicx}
\usepackage{subcaption}
\usepackage{pifont}
\usepackage{tabularx}

\title{Overview of ImageArg-2023: The First Shared Task in Multimodal Argument Mining}

\author{Zhexiong Liu,  Mohamed Elaraby\thanks{\;\;These authors contributed equally to this work.} , Yang Zhong\footnotemark[1]  ,  Diane Litman \\
        Department of Computer Science\\
        University of Pittsburgh,  Pittsburgh, PA 15260 USA \\ 
         \texttt{\{zhexiong.liu, mse30, yaz118, dlitman\}@pitt.edu} \\}

\begin{document}

\maketitle

\begin{abstract}
This paper presents an overview of the \textit{ImageArg} shared task, the first multimodal Argument Mining shared task co-located with the $10^{th}$ Workshop on Argument Mining at EMNLP 2023. The shared task comprises two classification subtasks - (1) Subtask-A: Argument Stance Classification; (2) Subtask-B:  Image Persuasiveness Classification. The former determines the stance of a tweet containing an image and a piece of text toward a controversial topic (e.g., gun control and abortion). The latter determines whether the image makes the tweet text more persuasive. The shared task received 31 submissions for Subtask-A and 21 submissions for Subtask-B from 9 different teams across 6 countries. The top submission in Subtask-A achieved an F1-score of $0.8647$ while the best submission in Subtask-B achieved an F1-score of $0.5561$.

\end{abstract}

\input{Sections/1.Introduction}
\input{Sections/2.RelatedWork}
\input{Sections/3.TaskDetails}
\input{Sections/4.Submission}
\input{Sections/5.Conclusion}
\input{Sections/6.Limitations}
\bibliography{anthology,custom}
\bibliographystyle{acl_natbib}
\appendix
\section{Appendix}
\label{sec:appendix}
\begin{table*}[]
    \centering
    \footnotesize
    \begin{tabularx}{\textwidth}{c|p{4cm}X}
    \toprule
       \textbf{Image} & \textbf{Text} & Annotations\\ 

        \midrule 
         \multirow{4}{*}{\includegraphics[height=.15\textwidth,width=3.3cm]{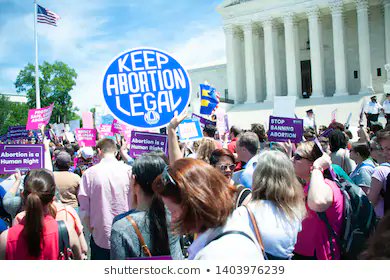}} &'Abortion law is pro-life. It saves 'mother over 'growing fetus in unwanted pregnancy due to rape, psychological trauma, social stigma, etc.  It stops back-alley abortions that kill. Counseling \& transition homes can lessen 'need for abortion. & \textbf{Topic:} Abortion \hfill \break \textbf{Annotated Label:}  Oppose \hfill \break \textbf{System Predictions:} \hfill \break \{'Oppose': 19, 'Support':12\} \hfill \break
         \textbf{Potentially Correct  Label: } Support 
          \hfill \break \textbf{Rationale: } The human annotation is inaccurate, super interesting on the usage of 'pro-life', to advocate for abortion.\\
          \midrule
          \multirow{4}{*}{\includegraphics[height=.15\textwidth,width=3.3cm]{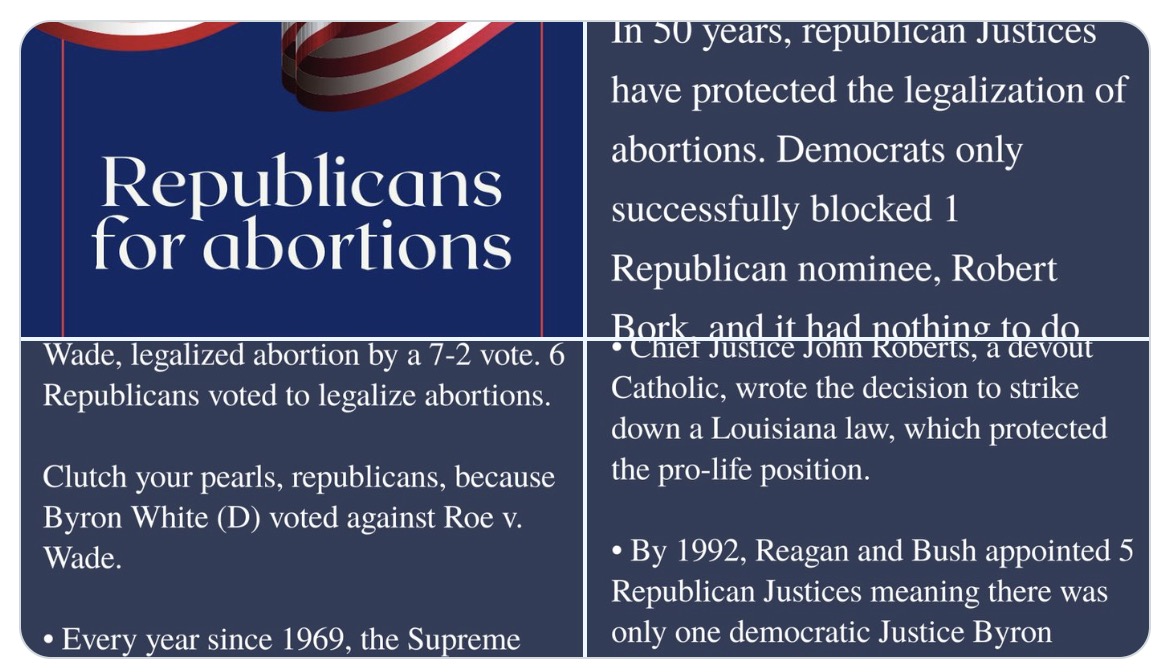}} & How Pro-Life is the Republican party and Justices? Facts matter here the answer, they're not. Thanks to their rulings, women have been able to safely have abortions. \#RoeVWade \#Republicans \#SCOTUShearings \#Constitution \#prochoice \#ProLife \#Facts & \textbf{Topic:} Abortion \hfill \break \textbf{Annotated Label:}  Support \hfill \break \textbf{System Predictions:} \hfill \break \{'Oppose': 20, 'Support':11\} \hfill \break
         \textbf{Potentially Correct  Label: } Support 
          \hfill \break \textbf{Rationale: } This tweet uses sarcasm, and is hard to annotate (republicans are in general not supporting legal abortion). Here the contents are image-dependent.\\
          
        \bottomrule
    \end{tabularx}
    \caption{Manually checked data with controversial scenarios for Subtask-A, where nearly half of the systems failed to predict the correct label. We sampled a few tweets and provided a potential correct label based on our manual inspections. The first example redefines a widely used anti-abortion term, pro-life, and advocates for abortion instead. The second is a complicated one that requires the comprehension of texts embedded in the image.  }
    \label{tab:manual_analysis}
\end{table*}

\end{document}

%% file: Sections/1.Introduction.tex
\section{Introduction}
Research in Argument Mining (AM) typically centers around the examination of an author's argumentative position, achieved through the automated identification of argument structures. This research has predominantly concentrated on domains presented in textual formats, encompassing endeavors such as mining persuasiveness in essays~\cite{stab2014annotating} and user-generated web discourse~\cite{habernal2017argumentation}. Recently, there has been a growing recognition of the need for multimodality in AM research. A noteworthy development in this regard is the \textit{Retrieval for Argument} shared task~\cite{carnot2023stance}. This task is designed to retrieve images related to a controversial topic that aligns with the textual stance, whether it supports or contradicts the topic. In a related context,~\citet{liu-etal-2022-imagearg} introduced the \textit{ImageArg} corpus, which is designed to investigate multimodal persuasiveness within tweets. This corpus represented an advancement in the field of automated persuasive text identification~\cite{duthie2016mining} by introducing a new modality through the inclusion of images.  

\begin{figure}[t]
    \centering
    \includegraphics[width=0.97\columnwidth]{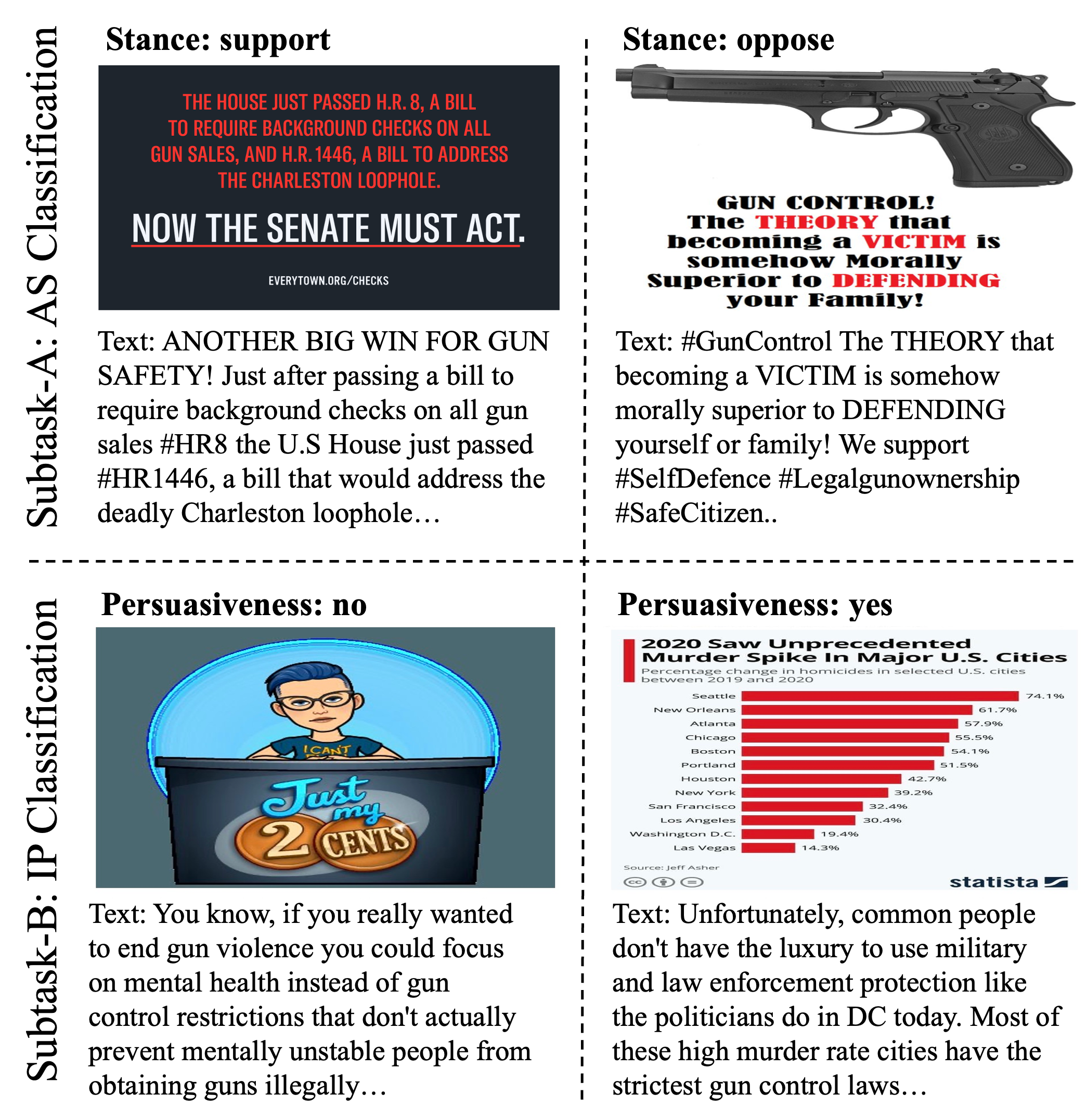}
    \caption{Examples of Subtask-A: Argument Stance (AS) Classification and Subtask-B: Image Persuasiveness (IP) Classification.}
    \label{fig:image-persuasiveness}
\end{figure}

This paper introduces the \textit{ImageArg} shared task\footnote{\url{https://imagearg.github.io/}},
building upon the groundwork laid by~\citet{liu-etal-2022-imagearg} and conducted as a part of the $10^{th}$ Workshop on Argument Mining\footnote{\url{https://argmining-org.github.io/2023/}}. The shared task comprises two subtasks that center around two highly controversial topics (gun control and abortion):
\begin{itemize}
\item 
Subtask-A: Argument Stance (AS) Classification. The primary objective is to determine, for each of these topics, whether a given tweet text and its accompanying image express either support or opposition. This subtask addresses the research question: how to identify an argument stance of the tweet that contains a piece of text and an image?
\item
Subtask-B: Image Persuasiveness (IP) Classification. The goal is to assess whether the image associated with a tweet makes the tweet text more persuasive or not. This subtask addresses the research question: does the tweet image make the tweet text more persuasive? 
\end{itemize}
Figure~\ref{fig:image-persuasiveness} shows examples of the two subtasks. The upper left tweet expresses a strong stance towards supporting gun control by indicating a house bill about the requirement of background checks for all gun sales. The upper right tweet opposes gun control because it is inclined to self-defense. The lower left tweet has an image irrelevant to the gun control topic. It does not improve the persuasiveness of the tweet text (and its stance) that argues to focus on mental health instead of gun restriction. The lower right tweet makes the tweet text (and its stance) more persuasive because it provides strong evidence to show the statistics of the murder rate in major U.S. cities due to restrictive gun control laws, so citizens cannot easily arm themselves.

The shared task received 31 submissions for Subtask-A and 21 submissions for Subtask-B from $9$ diverse teams, comprising both academic experts from various universities and industry researchers, across $6$ different countries. In general, the submissions that utilized text information from tweet images and performed data augmentation yielded favorable results for Subtask-A. The submissions that utilized unified multimodal models also achieved good performance in Subtask-B. The highest Subtask-A F1-score of $0.8647$ was attained by \textbf{Team KnowComp}~\cite{zong-etal-2023}, while the leading Subtask-B F1-score of $0.5561$ was attained by \textbf{Team feeds}~\cite{torky-etal-2023}. Details about task submissions are described in Section \ref{sec:submission}.  

%% file: Sections/2.RelatedWork.tex
\section{Related Work}
\paragraph{Multimodal Learning}
Recently, there has been increasing attention to assessing the ability of artificial intelligence models to process and understand multimodal input signals that occur in real-world applications~\cite{Zhang_2018_Equal, alwassel2020self}. In the vision-language domain, tasks are primarily designed to evaluate the capacity of models to comprehend visual data and articulate reasoning in language~\cite{goyal2017making, hudson2019gqa}. In addition,~\citet{Zheng_2021} are interested in the discourse relations between text and its associated images in recipes, while~\citet{kruk2019integrating} explores the multimodal document intent of Instagram posts. More recently,~\citet{liu-etal-2022-imagearg} introduce \textit{ImageArg}, the first multimodal learning corpus for argument mining. However,  the size of the \textit{ImageArg} corpus is small, which motivates our construction of an extension of the original corpus. Regarding multimodal modeling, researchers have developed methods to derive strong representations for each modality and implement fusion techniques~\cite{tsai2018learning, hu2019dense, tan2019lxmert, lu202012}. Although several shared tasks in machine translation~\cite{specia-etal-2016-shared, barrault-etal-2018-findings} and argument retrieval~\cite{carnot2023stance} have revealed the effectiveness of multimodal learning, none of them focused on argument persuasiveness. Therefore, this shared task provides opportunities to benchmark the new multimodal argument persuasiveness corpus by utilizing various image and text encoders along with effective fusion strategies. 

\paragraph{Computational Persuasiveness} 
While classical argument mining primarily focuses on the identification of argumentative components and their corresponding relationships~\cite{stab2014argumentation,stab2018cross, lawrence2020argument}, researchers have also focused on argument persuasiveness~\cite{chatterjee2014verbal, park2014computational, lukin2017argument, carlile2018give, chakrabarty2019ampersand}. Furthermore, while~\citet{riley1954communication},~\citet{o2015persuasion}, and~\citet{wei2016post} investigated the ranking of debate arguments on the same topic, they did not focus on discovering factors contributing to the persuasiveness of these arguments. In addition,~\citet{lukin2017argument} and~\citet{persing2017can} investigate how audience personality influences persuasiveness through diverse argument styles, such as factual versus emotional arguments. However, their work only focuses on the textual modality. In contrast,~\citet{higgins2012ethos} and~\citet{carlile2018give} focus their attention on persuasion strategies, e.g., Ethos (credibility), Logos (reason), and Pathos (emotion), within the context of reports and student essays. Building upon their work designed for textual corpora,~\citet{liu-etal-2022-imagearg} extend the annotation schemes to include the image modality. Although~\citet{park2014computational},~\citet{joo2014visual}, and~\citet{Huang2016InferringVP} employ facial expressions and bodily gestures to analyze persuasiveness within the realm of social multimedia, their investigations remain limited to human portraits and fail to generalize across diverse image domains. While prior work does explore persuasive advertisements in a multimodal fashion~\cite{Hussain_2017_Automatic, guo2021detecting}, it is important to note that their focus is on sentiment analysis, intent reasoning, and persuasive strategies tailored specifically for advertisements. In contrast, our shared task is interested in argument mining, marking an aligned goal to the \textit{ImageArg} work~\cite{liu-etal-2022-imagearg}, offering substantial value to multimodal computational social science.

%% file: Sections/3.TaskDetails.tex
\section{Corpus}
\begin{table}[t]
\centering
\small
\begin{tabular}{c|cc}
\toprule
Confidence         & Abortion & Gun control \\ \midrule
\textgreater{}= L5 & 0.8437    & 0.7434       \\
\textgreater{}= L4   & 0.7842    & 0.6697       \\
\textgreater{}= L3   & 0.7824    & 0.6551       \\
\textgreater{}= L2   & 0.7820    & 0.6516       \\
\textgreater{}= L1   & 0.7807    & 0.6487      \\ \bottomrule
\end{tabular}
\caption{Krippendorff's alpha for abortion and gun control topics with respect to different confidence levels.}
\label{tab:confidence}
\vspace{-1em}
\end{table}

We extended the \textit{ImageArg} corpus~\cite{liu-etal-2022-imagearg} by following its annotation protocol to annotate new data on abortion and gun control topics. Specifically, we annotated 1141 new abortion tweets and 301 new gun control tweets. Parts of the new gun control tweets were used to replace 131 out of the original 1003 gun control tweets in the \textit{ImageArg} corpus which were no longer available due to deletions or account suspensions. The other extras were annotated to ensure gun control and abortion tweets have close data distributions. Therefore, we obtained 1173 gun control tweets in total. In addition to using the original annotation protocol~\cite{liu-etal-2022-imagearg}, we required annotators to score confidence levels, which was designed to improve the inter-annotation agreement. Confidence was divided into 5 levels: L5-Extremely confident (understood and answered all annotations carefully), L4-Quite confident (tried to understand and answered most annotations carefully), L3-Somewhat confident (confused about some annotations), L2-Not very confident (did not understand some annotations), and L1-Not confident (mostly educated guesses).

In the annotation process, each tweet was annotated by three annotators on Amazon Mechanical Turk (AMT)\footnote{\url{https://www.mturk.com/}} who had done more than 5,000 approved annotations with at least 95\% approved rates in their historical hits. Annotators were required to pass a qualification exam that annotated pilot examples with at least 0.7 accuracy. Table~\ref{tab:confidence} shows AS annotation agreements in terms of Krippendorff's alpha~\cite{krippendorff2011computing} and confidence levels. We observed that annotations with high confidence levels had high agreements but dropped more annotations. To make the trade-off between annotation costs and agreements, we disregarded annotations with confidence levels less than L4 for abortion and less than L5 for gun control. The remaining new AS annotations for abortion and gun control have alpha scores of 0.78 and 0.74, respectively. The new IP annotations were also inherited from the \textit{ImageArg} protocol. First, annotators annotated two persuasiveness scores:
one for tweet text ($s_t$), another for tweet text and image ($s_{it}$). Then we computed a score difference $\Delta s_i = max(s_{it} - s_t, 0)$ as a persuasiveness gain from adding a tweet image. The final image persuasiveness score for each tweet was the average of persuasiveness gains from three annotators. To interpret image persuasiveness, we used the same threshold (0.5) in \textit{ImageArg} to split them into binary labels, indicating whether the image made the tweet text more persuasive or not. 

\begin{table}[t]
\centering
\footnotesize
\begin{tabular}{c|c|cccc|c}
\toprule
\multicolumn{1}{c|}{\multirow{2}{*}{Topic}} & \multicolumn{1}{c|}{\multirow{2}{*}{Split}} & \multicolumn{2}{c}{AS} & \multicolumn{2}{c|}{IP} & \multicolumn{1}{c}{\multirow{2}{*}{Total}} \\ \cmidrule{3-6}
\multicolumn{1}{c|}{}                       & \multicolumn{1}{c|}{}                       & Sup.    & Opp.    & Yes        & No         & \multicolumn{1}{c}{}                       \\ \midrule
\multirow{3}{*}{\begin{tabular}[c]{@{}c@{}}Gun \\ control\end{tabular}}                & train                                       & 475        & 448       & 251        & 672        & 923                                        \\
                                            & dev                                         & 54         & 46        & 33         & 67         & 100                                        \\
                                            & test                                        & 85         & 65        & 53         & 97         & 150                                        \\ \midrule
\multirow{3}{*}{Abortion}                   & train                                       & 244        & 647       & 278        & 613        & 891                                        \\
                                            & dev                                         & 19         & 81        & 26         & 74         & 100                                        \\
                                            & test                                        & 33         & 117       & 53         & 97         & 150                                        \\ \bottomrule
\end{tabular}
\caption{The data statistics for Subtask-A and Subtask-B for gun control and abortion topics.}
\label{tab:stat_taskA}
\vspace{-1em}
\end{table}

We split the corpus into train, development, and test sets in the shared task, which obtained 1814 train, 200 development, and 300 test samples for both subtasks\footnote{We removed one abortion tweet in the test set when we evaluated team submissions for the leaderboard because the tweet was no longer available during the task submission phase so a few teams were unable to download the full 300 test samples.}. The data
statistics are shown in Table~\ref{tab:stat_taskA} for Subtask-A and Subtask-B, respectively. We released the train and development data splits for model development and the test set without labels before the task submission deadline. We shared the complete test set with labels after completing the shared task. The full corpus can be downloaded from the GitHub repository\footnote{\url{https://github.com/ImageArg/ImageArg-Shared-Task}}.

%% file: Sections/4.Submission.tex
\begin{table*}[]
\footnotesize
\centering

\begin{tabular}{c|c|c|c|p{5.5em}|p{17em}}
\toprule
\renewcommand{\arraystretch}{0.6}
\textbf{ID} & \textbf{System} & \textbf{Score} & \textbf{Modality} & \textbf{Model} & \textbf{Notes}                                    \\\midrule
\textbf{1*}                             & KnowComp-4          & 0.8647         & I+T                                   & ResNet50 + DeBERTa                 & Augment Text with Back Translation + WordNet                    \\\midrule
2                              & KnowComp-5           & 0.8571         & I+T                                   & ResNet50 + DeBERTa                 & Augment Text with Translation + WordNet + Semantic SimilarityAttention \\\midrule
3                             & KnowComp-1           & 0.8528         & I+T                                   & ResNet101 + DeBERTa                      & Augment Text with Translation + WordNet                               \\\midrule
\textbf{4*}                              & Semantists-4         & 0.8506         & T+E                                 &                                    & Ensemble of All Models                                                \\\midrule
5                             & Semantists-3         & 0.8462         & T+E                                 & BERTweet                             & OCR on Image                                  \\\midrule
6                              & Semantists-5         & 0.8417         & T+E                                 & BERT                          & Dual Contrastive Loss + OCR on Image                                                          \\\midrule
7                             & Semantists-1        & 0.8365         &                     T+E                  &              BERT                       &   Contrastive Loss + OCR on Image                                                                \\\midrule
8                            & Semantists-2        & 0.8365         & T+E                                 & T5                                 & OCR on Image                                                          \\\midrule
9                             & KnowComp-2           & 0.8365         & I+T                                   & ResNet50 + DeBERTa                 & Augment Text with Translation + WordNet + Semantic SimilarityAttention                                          \\\midrule
10                             & KnowComp-3          & 0.8346         & I+T                                   & LayoutLMv3 + DeBERTa                  & Augment Text with Translation + WordNet                               \\\midrule
\textbf{11*}                             & Mohammad   Soltani-2 & 0.8273         & I+T                                   & CLIP32                             & AdaBoost for Abortion + Xgboost for Gun Control                       \\\midrule
\textbf{12*}                             & Pitt Pixel Persuaders-2      & 0.8168         & T                                     &                                    & Emsemble All The Model                                                \\\midrule
13                             & Mohammad   Soltani-1 & 0.8142         & I+T                                   & CLIP32                            & AdaBoost for Abortion and Gun Control                                 \\\midrule
14                             & Mohammad   Soltani-4 & 0.8093         & I+T                                   & CLIP32                             & Xgboost for Abortion and Gun   Control                                \\\midrule
\textbf{15*}                           & GC-HUNTER-2          & 0.8049         & T                                     & XLMRoberta                         &                                                                       \\\midrule
16                              & Mohammad   Soltani-3& 0.8000           & I+T                                   & CLIP32                             & AdaBoost for Abortion + RUSBoost for Gun Control                      \\\midrule
17                              & Pitt Pixel Persuaders-1       & 0.7910          & T                                     & BLOOM-560m                         &                                                                       \\\midrule
18                              & Mohammad   Soltani-5 & 0.7782         & I+T                                   & CLIP32                             & SVM-Poly for Abortion and Gun Control                                 \\\midrule
19                             & GC-HUNTER-1         & 0.7766         & T                                     & BERT                               &                                                                       \\\midrule
\textbf{20*}                             & IUST-1              & 0.7754         & T+E                                   & BERTweet                           & Augment  Text with ChatGPT   paraphraser + OCR on image               \\\midrule
21                              & IUST-2              & 0.7752         & T+E                                   & RoBERTa                            & Augment  Text with ChatGPT   paraphraser + OCR on image               \\\midrule
22                              & Pitt Pixel Persuaders-4        & 0.7710          & T                                     &         Bloom-1B                           &                                                                       \\\midrule
23                              & Pitt Pixel Persuaders-5     & 0.7415         & T                                     & XLNet                              &                                                                       \\\midrule
\textbf{24*}                              & KPAS-1               & 0.7097         & I+T                                   & CLIP                               &                                                                       \\\midrule
\textbf{25*}                             & ACT-CS-4            & 0.6325         & I+T+E+C                               & ViT+BERT                           & Cross-Attention                                                       \\\midrule
26                               & ACT-CS-3             & 0.6178         & I+T+E                                 & ViT+BERT                           & Cross-Attention                                                       \\\midrule
27                               & ACT-CS-2             & 0.6116         & I+T                                   & ViT+BERT                           & Cross-Attention                                                       \\\midrule
28                               & ACT-CS-1            & 0.5863         & I+T                                   & ViT+BERT                           &  Simple Concatenation of features                                  \\\midrule
29                              & IUST-3               & 0.5680         & I+T+E                                 & CLIP+BERT                          & Augment  Text with ChatGPT   paraphraser + OCR on image               \\\midrule
30                               & Pitt Pixel Persuaders-3        & 0.5285         & I+T                                   & ViLT                               &                                                                       \\\midrule
\textbf{31*}                               & feeds-1$^{**}$         & 0.4418         & T                                     & BERT                               &                                                                       \\
\bottomrule
\end{tabular}
\caption{The Subtask-A submission results. The System column refers to the Team name and submission attempt number connected by "-". Each Team has at most five submissions. The scores are positive F1 scores. The T, I, E, and C represent text, image, extracted text from image, and image caption modality, respectively. Rows with \textbf{bold} ID and marked with * refer to the best system for each participating team. ** Team feeds submitted results for one topic by the submission deadline, so only partial results are evaluated.}\label{tab:subtaskA_res}
\end{table*}

\section{Submission Results}
\label{sec:submission}
We provide summaries about Subtask-A (Sec.~\ref{sec:subtask_A}) and Subtask-B (Sec.~\ref{sec:subtask_B}) submissions for all the teams. In cases where a team did not submit a description paper, we include their results and provide a brief description based on the survey completed by the team at the time of submission.

\begin{figure*}[]
    \centering
    \small
    \includegraphics[width=1\textwidth]{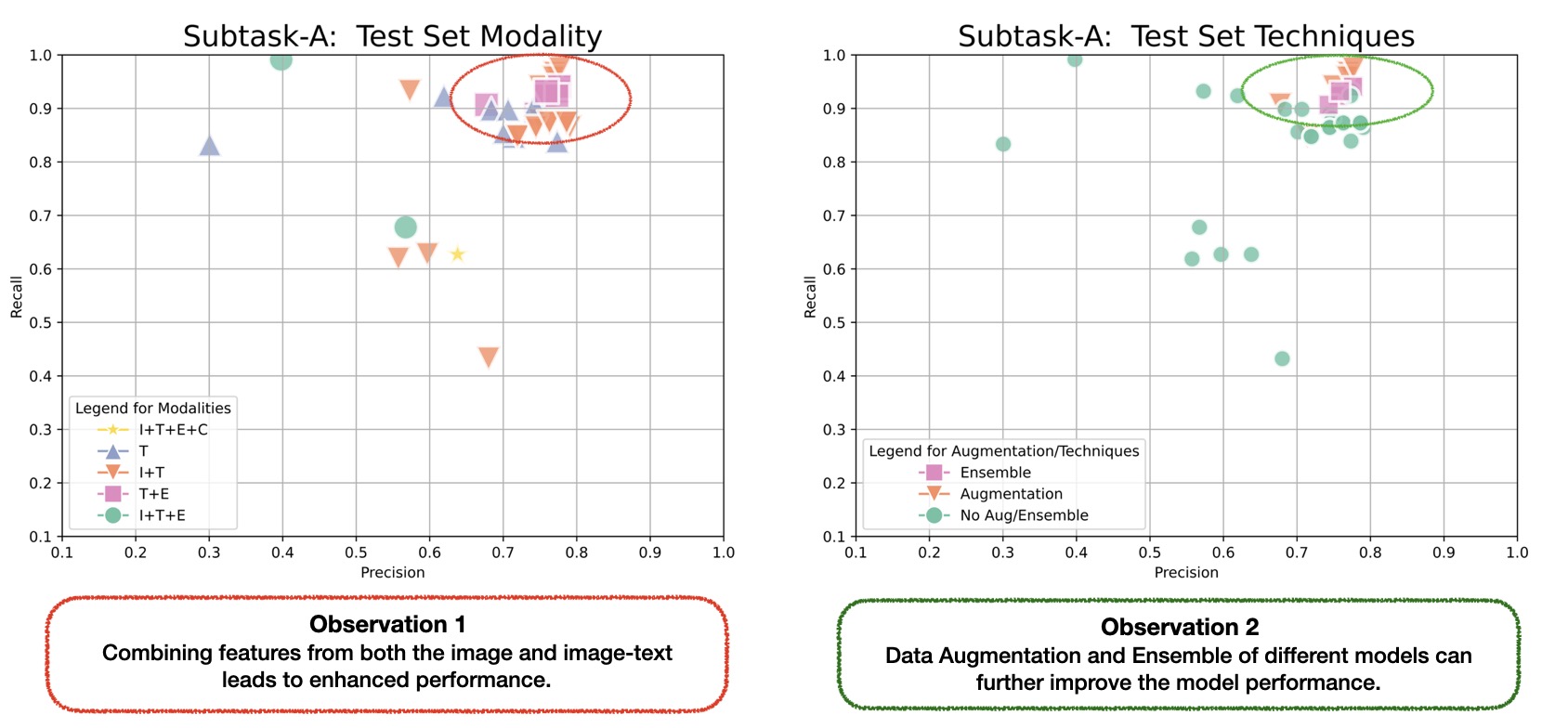}
    \caption{Subtask-A: system performance in relation to the computation approaches (left: modalities, right: techniques). We grouped systems based on the modalities used by the model (left) and computational techniques (right).  The T, I, E, and C represent text, image, extracted text from image, and image caption modality, respectively. }
    \label{fig:taskA_plot}
    \vspace{-1em}
\end{figure*}

\subsection{Subtask-A: AS Classification}
\label{sec:subtask_A}

Initially, we observed that models utilizing multimodal features (I+T or T+E) displayed higher performances, where I denotes tweet images, T denotes tweet text, and E denotes the text extracted from images. Table \ref{tab:subtaskA_res} illustrates that the top-performing submissions (top 10) employed two primary strategies: they either fused features extracted from both image and text encoders separately, or used pretrained language models finetuned on text extracted from images and tweets, which gave an additional textual context to the original tweet. This innovative method improved model performance compared to the ones that only used tweet text data in general\footnote{Results may vary depending on the model training details and experimental setups across participating teams}. Also, the last column shows that data augmentation exhibited promise, given the limited annotated data in this shared task.

\subsubsection{System Descriptions}
We describe representative methods from leading teams while summarizing the approaches from the remaining teams as follows:

\textbf{Team KnowComp} introduced a unified Framework for Text, Image, and Layout Fusion in Argument Mining, TILFA~\cite{zong-etal-2023}. They highlighted the need for better image encoding with textual information. To tackle the problem of unbalanced data, they augmented the tweet texts with backtranslation and synonym replacements. 

\textbf{Team Semantists}~\cite{rajaraman-etal-2023} submitted five system runs for task A, focusing mainly on the text-based approaches. To harness the information from the images, they extract text from the tweet image through an OCR system and concatenate it with the tweet texts. Pretrained language models such as  T5 NLI~\cite{2020t5} and BERTTweet are applied for label predictions. The team also adopts a Multi-task Contrastive Learning Framework similar to~\citet{chen2022dual} with the label aware augmentation for contrastive learning.

\textbf{Team Mohammad Soltani}~\cite{soltani-etal-2023} experimented with CLIP~\cite{RadfordKHRGASAM_21_CLIP} to extract the textual and visual modality features. They then combined features from both modalities by concatenating them along the last dimension according to an early fusion strategy, followed by traditional machine learning classifiers such as AdaBoostClassifier and SVM-Poly.

\textbf{Team Pitt Pixels Persuaders}~\cite{arushi-etal-2023} fine-tuned multiple text-based pre-trained models such as XLNet~\cite{yang2019xlnet} and BLOOM~\cite{scao2022bloom} on the corpus.~\textbf{Team IUST}~\cite{nobakhtian-etal-2023} did data augmentation using GPT to paraphrase tweet text and extracted text from images and finetuned text-based models.~\textbf{Team feeds}~\cite{torky-etal-2023} and \textbf{Team GC-Hunter}~\cite{shokri-etal-2023} only finetuned pre-trained language models on the tweet text. Both \textbf{Team ACT-CS}~\cite{zhang-etal-2023} and ~\textbf{Team KPAS} studied multimodal feature fusions.

\subsubsection{Method Discussions}
Table \ref{tab:subtaskA_res} reveals that the most successful submissions utilized pretrained language models such as DeBERTa, BERT, and BERTweet~\cite{bertweet}. Furthermore, the integration of data augmentation techniques, such as backtranslation and word substitution using WordNet, was observed to enhance performance, as depicted in Figure \ref{fig:taskA_plot}. This boost in performance can be attributed to the inherent reliance on textual information in the stance detection task. Augmenting the relatively limited annotated corpus with these techniques appears to be advantageous. Additionally, leveraging features from the visual modality, whether through image representations or image-text representations, further improved performance, ultimately leading to the highest overall scores, as demonstrated in Table \ref{tab:subtaskA_res} (rows 1 to 10).

On the other hand, the methods that utilized multimodal techniques like CLIP performed relatively lower than those that employed separate encoders for text and visual modalities. This is evident when referencing Table \ref{tab:subtaskA_res}, where the system achieving the highest performance using CLIP as the joint encoder, namely the submission by Mohammad Soltani-2, is ranked $11^{th}$ on the leaderboard. Additionally, it's noteworthy that only a limited number of teams explored the use of Large Language Models (LLMs). This might be attributed to our initial guidelines\footnote{\url{https://imagearg.github.io/}}, which indicated that the utilization of commercial APIs like chatGPT\footnote{\url{https://platform.openai.com/docs/guides/gpt/chat-completions-api}} would not contribute to the final ranking. Nevertheless, submissions that leveraged open-source LLMs, such as BLOOM-1B (row 22), exhibited lower performance compared to other submissions using pretrained language models. This opens up opportunities for further research into exploring the capabilities of LLMs in understanding argumentation, especially in multimodal contexts.

\subsubsection{Error Analysis}
\label{sec:error_analysis}
\begin{figure}
    \centering
    \includegraphics[width=1\columnwidth]{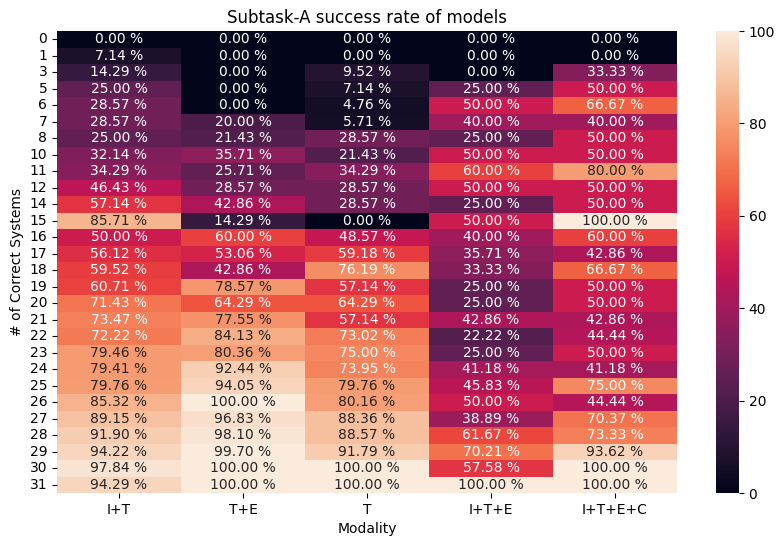}
    \caption{Average rate of correct predictions for Subtask-A systems (grouped by modalities) across tweet difficulties: the y-axis represents the number of systems making correct predictions out of 31 systems.}
    \label{fig:taskA_sucess_rate}
\end{figure}

Figure \ref{fig:taskA_sucess_rate} categorized the systems based on the modalities they incorporate and evaluated their respective success rates. Our analysis focused on system' ability to make accurate predictions, quantified by the number of successful systems out of 31 systems. We found that systems that incorporated both image and text modalities (I+T) generally yielded reasonable predictions, with at least one system in this category correctly identifying the label. Additionally, models that combined text and extracted text from images (T+E) displayed particularly strong performance, especially for data of intermediate difficulty. In these cases, the success rate for these systems exceeded 60\%, with at least 19 out of the 31 systems making correct predictions.

In a qualitative analysis of the 299 valid tweets in the test set, we found that 160 tweets (53\%) were accurately predicted by a majority of systems (>= 26 out of 31 systems). Among the subset of tweets (86) exhibiting intermediate difficulty (where 6-20 teams failed to predict the correct labels), we manually sampled ten tweets for label analysis and provided potentially correct labels.
Our findings indicate that these tweets often encompass cynicism or sarcasm regarding a specific topic (3 cases), are heavily reliant on the image contents/charts (3 cases), or can be traced back to annotation noise or contents unrelated to the provided topic. Detailed insights are shown in Table \ref{tab:manual_analysis} in Appendix~\ref{sec:appendix}. For instance, the first example associates "pro-life" with "Abortion Law", suggesting the tweets favor abortion. In the second example, a deep understanding of the text embedded within images is crucial for providing accurate labels. These observations underscore the complexities in multimodal argument mining tasks and highlight the critical role of cross-modal information fusion.

\subsection{Subtask-B: IP Classification}
\label{sec:subtask_B}

\begin{table*}[t]
\footnotesize
\centering
\begin{tabular}{c|c|c|c|p{5.5em}|p{17em}}
\toprule
\textbf{ID} & \textbf{System}      & \textbf{Scores} & \textbf{Modality} & \textbf{Model}               & \textbf{Notes}                                                   \\\midrule
\textbf{1}*           & feeds-1              & 0.5561          & I+T               & CLIP                & Cleaned Text                    \\\midrule
\textbf{2}*           & KPAS-2               & 0.5417          & I+T               & CLIP                &                                                         \\\midrule
3           & feeds-2              & 0.5392          & I+T               & CLIP                & Uncleaned Text                                          \\\midrule
\textbf{4}*          & Mohammad   Soltani-5 & 0.5281          & I+T               & CLIP32+REL +Convnext &                                                         \\\midrule
\textbf{5}*           & Semantists-1         & 0.5045          & T+E               & T5                  & OCR on Image                                            \\\midrule
\textbf{6}*           & ACT-CS-1             & 0.5000             & I+T               & Vit+BERT            &                                                         \\\midrule
7           & Mohammad   Soltani-1 & 0.4875          & I+T               & CLIP32              & SVM-Poly for Abortion LogisticReg for Gun Control       \\\midrule
8           & Mohammad   Soltani-4 & 0.4778          & I+T               & CLIP32+REL +Convnext & SGD for Abortion LogisticReg for Gun Control            \\\midrule
9           & Mohammad   Soltani-3 & 0.4762          & I+T               & CLIP\_L\_14         & SVM-Poly for Abortion and Gun Control                    \\\midrule
10          & Semantists-5         & 0.4659          & T+E               &                     & Emsemble with majority vote                             \\\midrule
\textbf{11}*          & IUST-1              & 0.4609          & I+T               & CLIP+BERT           & Augment  Text with   ChatGPT paraphraser + OCR on image \\\midrule
12          & Mohammad   Soltani-2 & 0.4545          & I+T               & CLIP32              & SGD for Abortion and Gun Control                        \\\midrule
13          & ACT-CS-4             & 0.4432          & I+T+E+C           & Vit+BERT            & Cross Attention                                         \\\midrule
14          & ACT-CS-3             & 0.4348          & I+T+E             & Vit+BERT            & Cross Attention                                         \\\midrule
15          & Semantists-4        & 0.4222          & T+E               &                     & Emsemble with consistency loss                          \\\midrule
16          & Semantists-2         & 0.4141          & T+E               & Stancy BERT         &                                                         \\\midrule
\textbf{17}*          & KnowComp-1          & 0.3922          & I+T               & LayoutLMv3 +DeBERTa  & Augment Text with Translation + WordNet                 \\\midrule
\textbf{18}*          & GC-HUNTER-1          & 0.3832          & I+T+E             & ViLT                & OCR on Image                                            \\\midrule
19          & ACT-CS-2             & 0.3125          & I+T               & Vit+BERT            & Cross Attention                                         \\\midrule
20          & Semantists-3         & 0.2838          & I+T+E             & ALBEF               &                                                        \\\midrule
\textbf{21}*          & Pitt Pixel Persuaders-1        & 0.1217          & I+T               & CLIP                &                                                         \\
\bottomrule
\end{tabular}
\caption{The Subtask-B submission results. Each Team is allowed at most 5 submissions. The scores are positive label F1. The T, I, E, and C, represent text, image, extracted text from image, and image caption modality, respectively. Rows with \textbf{bold} ID and marked with * refer to the best system for each participating team.}
\vspace{-1em}
\label{tab:subtaskb_res}
\end{table*}

In contrast to Subtask-A, participating teams made fewer submission attempts for Subtask-B (a total of $21$ compared to $31$ for Subtask-A). Notably, all submissions in Subtask-B employed approaches that incorporated multiple modalities, as this task inherently requires an integration of visual and textual information to assess image persuasiveness.

As shown in Table \ref{tab:subtaskb_res},  utilizing CLIP~\cite{RadfordKHRGASAM_21_CLIP} model is evident to be the most effective technique in extracting multimodal features, which yields the best results (top-4 systems leveraged CLIP). This indicates that a unified encoder can better model the cross-modal information fusion, compared to employing individual models (i.e., ViT~\cite{dosovitskiy2020image} for image and BERT~\cite{devlin2019bert} for text) for feature extractions. Moreover, three teams utilized off-the-shelf Optical Character Recognition (OCR) tools to extract image text content. This extracted text was then combined with the original tweet texts to fine-tune pre-trained language models, which suggests that users could include arguments through texts embedded in the images.

\begin{figure*}[t]
    \centering
    \small
    \includegraphics[width=1\textwidth]{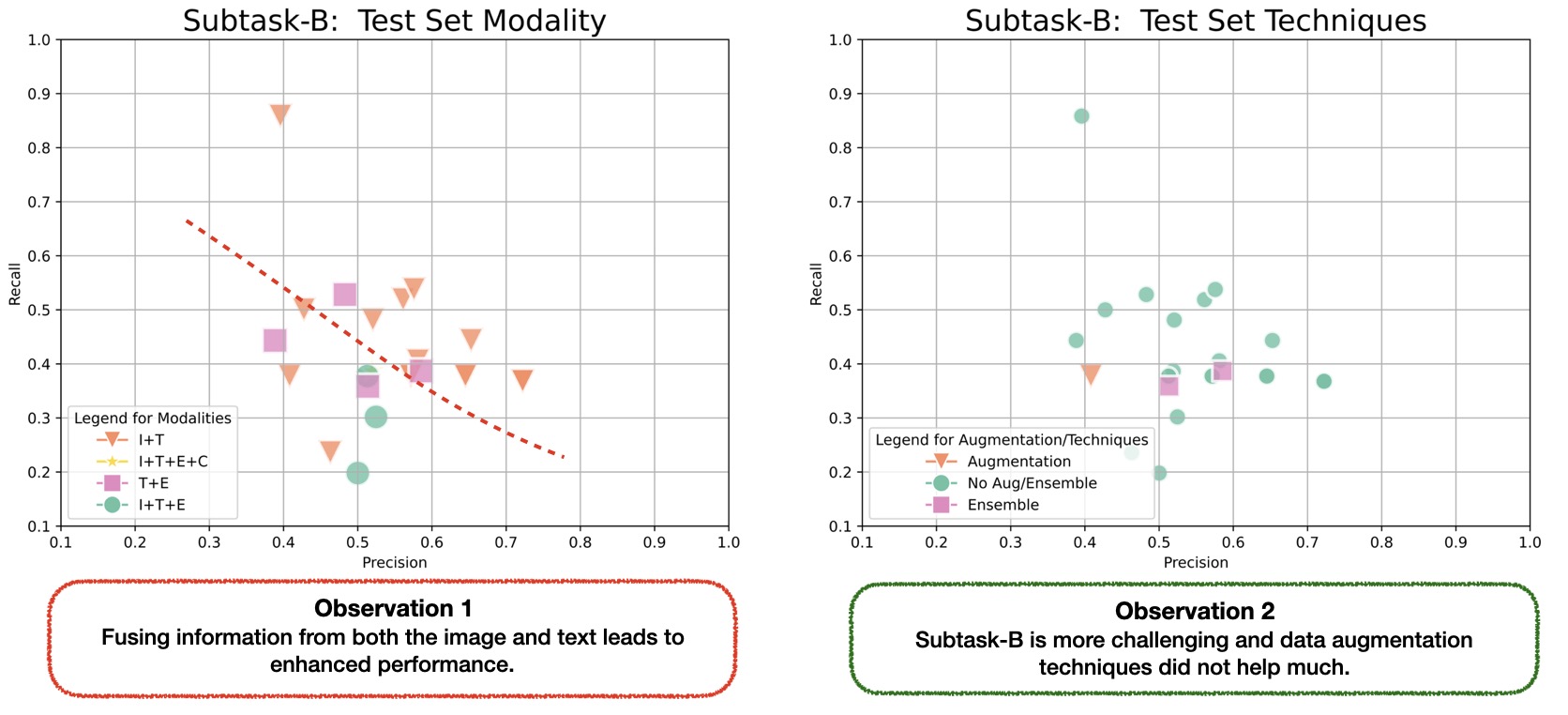}
    \caption{Subtask-B: system performance in relation to the computation approaches (left: modalities, right: techniques). We grouped systems based on the modalities used by the model (left) and computational techniques (right).  The T, I, E, and C represent text, image, extracted text from image, and image caption modality, respectively. }
    \label{fig:taskb_plot}
\end{figure*}

\subsubsection{System Descriptions}
We describe systems from the top-performing teams and briefly summarize the remaining teams:\footnote{While Team KPAS was among the top-performing teams, they did not submit a system description paper.}

\textbf{Team feeds}~\cite{torky-etal-2023} made $2$ submissions (Table \ref{tab:subtaskb_res} rows $1$ and $3$). The team utilized the CLIP model to encode the image and text and use a simple concatenation to fuse the two modalities, then trained a neural network on the concatenated features. They carefully cleaned tweet texts by recovering common abbreviations with their full forms (such as "I'm to I am") and also removed content such as URLs, emails, and phone numbers.

\textbf{Team KPAS} did not submit a system demonstration paper. However, their submission notes showed that they also employed the CLIP model to extract multimodal features. 

\textbf{Team Mohammad Soltani}~\cite{soltani-etal-2023} made a total of $5$ submissions (Table \ref{tab:subtaskb_res} rows $4$, $7$, $8$, $9$, and $12$). Notably, they adopted a topic-specific approach, tailoring their strategies to each topic separately. For the "Abortion" topic, they integrated visual features extracted from the CLIP model and utilized them as inputs for a classifier. Conversely, when tackling the "gun control" topic, their most successful model was crafted by combining features from Reformer~\cite{kitaev2019reformer}, ELECTRA~\cite{clark2019electra}, and LayoutLM~\cite{layoutlm}.

Similar to the systems in Subtask-A, \textbf{Team Semantists}~\cite{rajaraman-etal-2023} extracted texts from images and fine-tuned pretrained Language models such as T5 NLI and StancyBERT~\cite{popat2019stancy} on the corpus. \textbf{Team ACT-CS}~\cite{zhang-etal-2023} and \textbf{Team KnowComp}~\cite{zong-etal-2023} used separate models to encode the visual and textual information individually, then fine-tuned classifiers based on the fused features. \textbf{Team IUST}~\cite{nobakhtian-etal-2023} (Table \ref{tab:subtaskb_res} row $11$) leveraged the {MultiModal Bit Transformer} to extract features from both image and text sources concurrently. \textbf{Team GC-Hunter}~\cite{shokri-etal-2023} chose to concatenate text content from both tweets and OCR outputs to fully leverage textual information, complemented by image features extracted from a separately trained {ViLT} model. Finally, \textbf{Team Pitt Pixel Persuaders}~\cite{arushi-etal-2023} (Table \ref{tab:subtaskb_res}, row $21$) did not include the details of their Subtask B submission in their system description paper. However, their submission notes reveal that they also relied on CLIP, which proved to be less successful in their case.

\subsubsection{Method Discussion}

Figure \ref{fig:taskb_plot} illustrates that, unlike Subtask A, the application of data augmentation techniques which primarily concentrated on augmenting the text modality exclusively obtained only modest improvements in classification performance. Notably, none of the participating teams explored augmentation for the visual modalities, which presents an opportunity for further research into the impact of image augmentation on enhancing persuasiveness detection.

Additionally, Table \ref{tab:subtaskb_res} indicates that none of the submissions integrated LLMs into their systems. This observation can also be attributed to the task's primary emphasis on both visual and textual modalities and the guidelines we enforced, which limited the use of LLMs to open-source models. These open-source models have received less attention within the context of multimodal tasks, providing an explanation for their absence in the submissions.

\subsubsection{Error Analysis}
Figure \ref{fig:taskB_success_rate} categorizes the systems based on the modalities they incorporate and their respective success rates. Our analysis focused on the models' ability to make accurate predictions, quantified by the number of successful systems out of the 21 total systems. We found that systems incorporating both image and text modalities (I+T) consistently produced accurate predictions across data points with varying levels of difficulty. Interestingly, systems that combined text, text on images, images, and captions (I+T+E+C) demonstrated strong performance, particularly for data with high difficulty levels (as indicated by rows where only 4/5 systems made correct predictions). As reported by~\citet{soltani-etal-2023}, these systems tended to classify images showing only text as persuasive. Further analysis on the data illustrated different argumentation techniques, such as cases, consequences, or outcomes related to the textual argument, further highlighting the complexity and diversity of approaches employed in this shared task.

%% file: Sections/5.Conclusion.tex
\section{Conclusion}
In this paper, we introduced the \textit{ImageArg} shared task, marking a significant milestone as the inaugural shared task in multimodal argument mining, co-located with the $10^{th}$ Argument Mining Workshop at EMNLP 2023. A total of 9 teams from 6 different countries enthusiastically participated in this task, collectively submitting $31$ systems for Subtask-A Argument Stance (AS) classification and $21$ systems for Subtask-B Image Persuasiveness (IP) classification. The results reveal that Subtask-A is comparatively more predictable than Subtask-B. Models that utilized both textual information and the text embedded within images demonstrated considerable performance in Subtask-A. Furthermore, the strategic use of data augmentation and ensemble methods further enhanced the models' effectiveness. In contrast, Subtask-B witnessed the predominant adoption of CLIP for feature extraction from both images and texts, a technique that exhibited significant promise. The two subtasks offered valuable opportunities for participants to actively engage and foster fruitful exchanges in multimodal argument mining research.

%% file: Sections/6.Limitations.tex
\section{Limitations}
In this section, we discuss the limitations of our work from multiple perspectives.
First, the datasets utilized in this task may not sufficiently cover a broad range of multimodal data, possibly leaning toward social media content related to two specific topics: gun control and abortion. The language of data included in the paper is English, which is limited and should be extended to other languages for argument mining. Meanwhile, as demonstrated in Section \ref{sec:error_analysis}, the label annotations may exhibit inconsistencies or inaccuracies, given the inherent complexity of the task. Also, the use of rhetorical devices, especially in addressing challenges like sarcasm detection, remains an underexplored area. The evaluation metrics employed may not fully encompass the nuanced performance aspects crucial for multimodal argument mining. Lastly, it's important to acknowledge that participating systems may encounter challenges when attempting to generalize their approaches across diverse data types, domains, or modalities.

\begin{figure}[t]
    \centering
    \includegraphics[width=1\columnwidth]{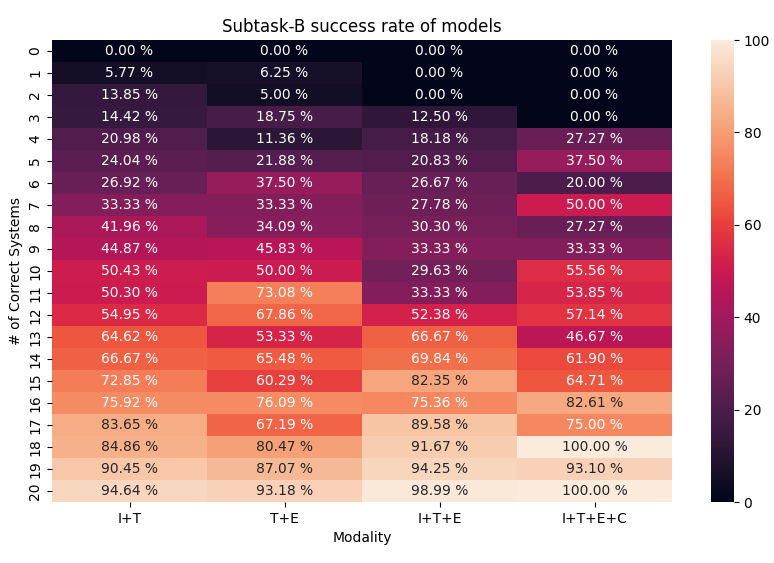}
    \caption{Average rate of correct predictions for Subtask-B systems (grouped by modalities) across tweet difficulties: the y-axis represents the number of systems making correct predictions out of 21 systems.}
    \label{fig:taskB_success_rate}
\end{figure}

Regarding the analysis of the results, it's important to acknowledge that since we mainly collected final predictions for both subtasks, the interpretability of the systems might remain unclear, presenting challenges in gaining insights into their decision-making processes. The intricate nature of multimodal argument mining can lead to multiple valid interpretations, potentially affecting the clarity of the ground truth.

\section{Ethics}
We acknowledge that there are privacy and ethical considerations in the collection and utilization of social media data. It's possible that biases within the dataset or system outputs may not have been fully mitigated. Given that our data originates from Twitter and the annotators predominantly come from English-speaking countries, it's inevitable that cultural biases are inherent in the data. However, we have implemented several measures to mitigate potential risks. To address privacy concerns, we have chosen to publicly share only the tweet IDs with the research community, which aligns with Twitter Developer Policy\footnote{\url{https://developer.twitter.com/en/developer-terms/policy}}.